\documentclass[conference]{IEEEtran}
\IEEEoverridecommandlockouts
\usepackage{cite}
\usepackage{amsmath,amssymb,amsfonts}
\usepackage{algorithmic}
\usepackage{graphicx}
\usepackage{textcomp}
\usepackage{xcolor}
\usepackage{todonotes}

\def\BibTeX{{\rm B\kern-.05em{\sc i\kern-.025em b}\kern-.08em
    T\kern-.1667em\lower.7ex\hbox{E}\kern-.125emX}}

\usepackage{comment}

\usepackage{array}  %
\usepackage{graphicx}  %
\usepackage{amsmath, bm}  %
\usepackage{amsthm}   %
\usepackage{amsfonts} %
\usepackage{amssymb} %
\usepackage{mathtools}%
\usepackage{glossaries}

\usepackage[ruled,vlined,linesnumbered]{algorithm2e}

\usepackage{subfig}
\usepackage[export]{adjustbox}

\usepackage[inline, shortlabels]{enumitem}

\usepackage{mathtools}

\usepackage{wrapfig}

\usepackage{hyperref}

\usepackage{dblfloatfix}

\renewcommand{\vec}[1]{\bm{#1}}
\newcommand{\mat}[1]{\mathbf{#1}}

\newcommand{\statespace}{\mathcal{S}}
\newcommand{\actionspace}{\mathcal{A}}

\newcommand{\state}{\vec{s}}
\newcommand{\nextstate}{\vec{s}'}

\newcommand{\action}{\vec{a}}

\newcommand{\policy}{\pi}
\newcommand{\policytheta}{\pi_{\theta}}

\newcommand{\thetavec}{\vec{\theta}}
\newcommand{\phivec}{\vec{\phi}}

\DeclarePairedDelimiterX{\infdivx}[2]{(}{)}{%
	#1\;\delimsize\|\;#2%
}

\newcommand{\R}{\mathbb{R}}
\newcommand{\E}[2]{\mathbb{E}_{#1}\left[#2\right]}

\newcommand{\transpose}{\intercal}

\newcommand{\Ptrans}{\mathcal{P}}

\newcommand{\policyparams}{\thetavec}

\newcommand{\gradpolicyparams}{\nabla_{\thetavec}}
\newcommand{\policyparametrized}{\policy_{\policyparams}}

\newcommand{\policynominal}{\pi_{\text{nom}}}

\newcommand{\trajectoryvec}{\vec{\tau}}
\newcommand{\BasisFunctionMatrix}{\mat{\Phi}}
\newcommand{\BasisFunctionVector}{\vec{\phi}}

\newcommand{\prompweights}{\vec{\omega}}
\newcommand{\meanprompweights}{\vec{\mu}_{\vec{\omega}}}
\newcommand{\covprompweights}{\vec{\Sigma}_{\vec{\omega}}}
\newcommand{\meanprompweightsconditioned}{\overline{\vec{\mu}}_{\prompweights}}
\newcommand{\covprompweightsconditioned}{\overline{\mat{\Sigma}}_{\prompweights}}

\newcommand{\position}{\vec{p}}
\newcommand{\deltaposition}{\vec{\delta}_{\position}}

\newcommand{\orientation}{\vec{q}}

\newcommand{\quaternion}{\vec{q}}
\newcommand{\deltaquaternion}{\vec{\delta}_{\quaternion}}

\newcommand{\torquevec}{\vec{\tau}}

\newcommand{\jointpositionvec}{\vec{q}}
\newcommand{\jointvelocityvec}{\dot{\vec{q}}}

\newcommand{\wrench}{\vec{\mathcal{F}}}

\newcommand{\interestintargetpose}{{}^{T}\mat{X}^{I}}

\newcommand{\eeintargetpose}{{}^{T}\mat{X}^{E}}
\newcommand{\eeintargetpos}{{}^{T}\position^{E}}
\newcommand{\eeintargetori}{{}^{T}\orientation^{E}}

\newcommand{\eeininterestpose}{{}^{I}\mat{X}^{E}}

\newcommand{\eepose}{\mat{X}^{E}}

\newcommand{\axisangle}{\vec{\nu}}
\newcommand{\deltaaxisangle}{\vec{\delta}_{\axisangle}}

\glsdisablehyper
\newacronym{ml}{ML}{Machine Learning}
\newacronym{rl}{RL}{Reinforcement Learning}
\newacronym{mdp}{MDP}{Markov Decision Process}

\newacronym{reptrick}{Rep-trick}{Reparametrization trick}
\newacronym{mvd}{MVD}{Measure-Valued Derivative}
\newacronym{sf}{SF}{Score-function}

\newacronym{vae}{VAE}{Variational Autoencoder}
\newacronym{vi}{VI}{Variational Inference}
\newacronym{elbo}{ELBO}{Evidence Lower Bound}

\newacronym{mc}{MC}{Monte Carlo}

\newacronym{fd}{FD}{Finite-difference}

\newacronym{lqr}{LQR}{Linear-Quadratic Regulator}

\newacronym{pgpe}{PGPE}{Policy Gradients with Parameter based Exploration}
\newacronym{nes}{NES}{Natural Evolutionary Strategies}
\newacronym{rwr}{RWR}{Reward Weighted Regression}
\newacronym{reps}{REPS}{Episodic Relative Entropy Policy Search}
\newacronym{nac}{NAC}{Natural Actor Critic}
\newacronym{power}{PoWER}{Policy Learning by Weighting Exploration with the Returns}
\newacronym{expected-sarsa}{Expected SARSA}{Expected SARSA}

\newacronym{trpo}{TRPO}{Trust Region Policy Optimization }
\newacronym{ppo}{PPO}{Proximal Policy Optimization}
\newacronym{ddpg}{DDPG}{Deep Deterministic Policy Gradient}
\newacronym{td3}{TD3}{Twin Delayed DDPG}
\newacronym{sac}{SAC}{Soft Actor-Critic}
\newacronym{gae}{GAE}{Generalized Advantage Estimation}

\newacronym{gpomdp}{}{GPOMPDP}
\newacronym{reinforce}{}{REINFORCE}

\newacronym{emvd}{E-MVD}{Episodic-MVD}
\newacronym{sacmvd}{SAC-MVD}{Soft Actor-Critic with MVD}
\newacronym{treemvdpg}{Tree-MVD}{Tree-MVD-Policy Gradient}

\newacronym{sacextrasamples}{SAC-extra-samples}{}

\newacronym{sacsf}{SAC-SF}{}
\newacronym{sacsfextrasamples}{SAC-SF-extra-samples}{}

\newacronym{spsa}{SPSA}{Simultaneous Perturbation Stochastic Approximation }

\newacronym{mp}{MP}{Movement Primitive}    
    
\begin{document}

\title{Residual Robot Learning for Object-Centric 
Probabilistic Movement Primitives \\

}

\author{\IEEEauthorblockN{Jo\~{a}o Carvalho, Dorothea Koert, Marek Daniv, Jan Peters}
\thanks{Authors are with the Intelligent Autonomous Systems,
Computer Science Department, Technische Universit\"{a}t Darmstadt, Germany, The work was funded by German Federal Ministry of Education and Research (project IKIDA, 01IS20045)

\{joao,doro,jan\}@robot-learning.de, \{marek.daniv\}@stud.tu-darmstadt.de}

}

\maketitle

\begin{abstract}
It is desirable for future robots to quickly learn new tasks and adapt learned skills to constantly changing environments. 
To this end, Probabilistic Movement Primitives (ProMPs) have shown to be a promising framework to learn generalizable trajectory generators from distributions over demonstrated trajectories. 
However, in practical applications that require high precision in the manipulation of objects, the accuracy of ProMPs is often insufficient, in particular when they are learned in cartesian space from external observations and executed with limited controller gains. 
Therefore, we propose to combine ProMPs with recently introduced Residual Reinforcement Learning (RRL), to account for both, corrections in position and orientation during task execution. 
In particular, we learn a residual on top of a nominal ProMP trajectory with Soft-Actor Critic and incorporate the variability in the demonstrations as a decision variable to reduce the search space for RRL.
As a proof of concept, we evaluate our proposed method on a $3$D block insertion task with a 7-DoF Franka Emika Panda robot.
Experimental results show that the robot successfully learns to complete the insertion which was not possible before with using basic ProMPs.
\end{abstract}


\section{Introduction}

The ability to learn new tasks from non-robotic experts would be of great benefit for future robots in industry, as well as everyday applications. To this end, Learning from Demonstration (LfD)~\cite{billard2008survey}, and particularly movement primitives, offer a way to learn generalizable trajectory generators, from a few demonstrated trajectories. 
However, in precise or contact-rich manipulation tasks, often pure imitation of demonstrated behaviors does not work well~\cite{kroemer2021review}.
Additionally, in real robotic applications controller accuracy might differ between robots or tasks, and lower controller gains, desirable e.g. to ensure safety in close cooperation with humans, could lead to tracking inaccuracies of trajectories learned from demonstrations. 

Recently introduced Residual Reinforcement Learning (RRL)~\cite{silver2018residual,johannink2019residual,schoettler2020deep} has shown success in combining classical controllers with residual policies learned through direct interactions with simulated or real environments.
In this paper, we explore the direction of combining RRL with movement primitives to use advantages of both in a symbiotic way.
This has already been investigated for the cases of Dynamic Movement Primitives (DMPs)~\cite{davchev2022residual} and Gaussian Mixture Model (GMM) based Primitives~\cite{nematollahi2021robot}.
Inspired by these works, we are investigating here the benefit of RRL for another movement primitive representation, namely Probabilistic Movement Primitives (ProMPs)~\cite{paraschos2013promp,paraschos2018using}.
ProMPs encode demonstrated trajectories as distributions over weighted basis functions and allow to capture variability and correlations within demonstrated trajectories.
Probabilistic operators such as conditioning on different goals and start points can then be used to generalize from demonstrations to unseen situations. 
While this works well for tasks demonstrated with joint-space kinesthetic teaching, e.g. assisting in coffee making~\cite{GomNeuSchPet20}, these conditioning operations often are too imprecise for insertion tasks, in particular when executed on a real robot with limited controller gains.

\begin{figure}[t]
  \centering
  \includegraphics[width=0.95\columnwidth]{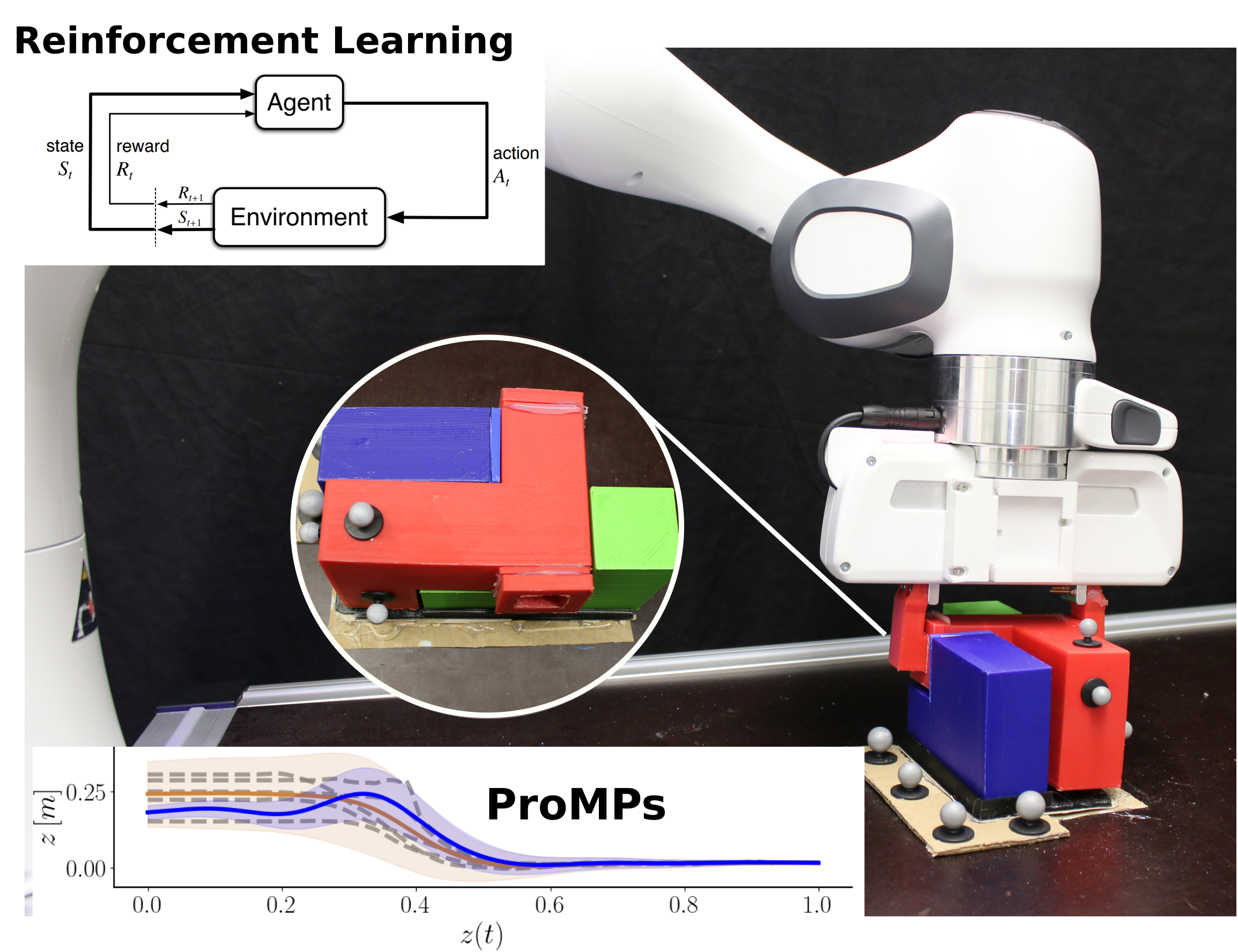}
  \caption{The robot learns to insert the red block from the Ubongo$3$D game with ProMPs adapted with residual reinforcement learning in cartesian space.
  The tolerance for insertion is approximately $3$mm.
  }
  \label{fig:real_robot_ubongo}
\end{figure}

To overcome these limitations, we propose to combine ProMPs with RRL, such that during task execution a robot can refine and iteratively improve trajectories. In particular, we investigate in an experimental study whether ProMPs, which were learned from external observations, can be used in combination with RRL to solve a precision $3$D block insertion task. The main contributions of this paper are hereby the following:
\begin{itemize}
    \item On top of a nominal trajectory generated with a ProMP we propose to learn a residual to account for both corrections in position and orientation with Soft-Actor Critic (SAC)~\cite{Haarnoja2018SAC} in a real robotic system.
    \item We use the variability in the demonstrations as a decision variable to reduce the search space for RRL and compare this approach to a distance-based strategy for weighting nominal and residual policy.
    \item We evaluate the proposed method on a $3$D block insertion task on a $7$-DoF Franka Emika Panda robotic arm. Contrary to peg-in-hole, this task is not invariant to rotations around all axes.
\end{itemize}

The rest of this paper is structured as follows.
In Section \ref{sec:related} we discuss related work for RRL with classical controllers and LfD.
In Section \ref{sec:approach} we introduce our approach to combine RRL with an object-centric formulation of ProMPs.
Here we also discuss three different options for the combination of the residual and nominal policies, which we then evaluate on a $3$D block insertion task with a real robot in Section \ref{sec:experiments}.
In Section \ref{sec:conclusion} we draw conclusions, discuss current limitations of the proposed approach and give an outlook on future work.

\section{Related Work}\label{sec:related}

Residual reinforcement learning~\cite{silver2018residual,johannink2019residual,schoettler2020deep} has been proposed as a way to solve challenging robotic manipulation tasks by adapting control actions from a conventional (model-based) controller with a policy learned with model-free RL, which significantly reduces the search space and thus improves sample efficiency.
\cite{silver2018residual} introduced RRL as learning a residual on top of an initial controller to improve non-differentiable policies using model-free deep RL.
\cite{johannink2019residual} uses a residual policy and a hand-engineered base controller in a robotic task to insert a block between two others that can topple, using a dense reward and a fined tuned vision setup.
\cite{alakuijala2021residual} extends RRL to learn from sparse rewards and visual input demonstrations as image sequences in simulation. Through behavioral cloning, they first learn task-relevant visual features.
Afterwards, the RL policy is optimized using the pretrained state features.
\cite{schoettler2020deep} applies deep RRL to industrial tasks in the real world, uses feature state from images, a sparse reward, and a hand-designed P-controller as a nominal policy.
The cartesian actions are transformed to joint-space via inverse kinematics.
\cite{kulkarni2021learning} learns assembly tasks with a real robot in a few minutes, combining cartesian impedance control with a recurrent policy version of TD3~\cite{fujimoto2018td3} to learn in the presence of position uncertainties.
Variable Impedance End-Effector Space (VICES)~\cite{martin2019variable} studied the effects of different action spaces, and argued for impedance control in end-effector space.
Besides learning a policy to learn small changes in pose, they also learn state-dependent gains for the impedance controller.
Older work~\cite{kim2009impedance} learned impedance gains of a robotic arm based on the equilibrium point control theory and the natural actor-critic algorithm~\cite{PetersNAC2008} for contact tasks. 
\cite{beltran2020learning} learns force control for rigid position-controlled robots and its follow up work~\cite{beltran2020variable} solves peg-in-hole tasks with hole-position uncertainty with an off-policy, model-free RL method using several sim2real techniques, such as domain randomization.
\cite{wang2021robotic} imitate human assembly skills through hybrid trajectory and force learning with hierarchical imitation learning.
A scheme to learn an optimal force control policy with goal conditioned imitation learning is presented in~\cite{ding2019goal} and closely connect to~\cite{wang2021hybrid}.
Guided Uncertainty-Aware Policy Optimization (GUAPO)~\cite{lee2020guided} quantifies uncertainty in pose estimation to define a binary switching strategy that determines where to use model-based or RL policies.
\cite{hoppe2020sample} evaluates Qgraph-bounded DDPG for improving model-free RL to solve a peg-in-hole with a force-torque action space.
\cite{ranjbar2021residual} proposes a hybrid RRL to modify the signals used by the RL policy to prevent internal feedback signals of the low-level controller limiting the RL agent to adequately improve its policy and thus harming learning. Their approach is shown in a contact-rich peg-insertion task.
\cite{shi2021combining} combines visual servoing-based LfD and force-based Learning by Exploration (LbE), and proposes region-limited residual RL (RRRL) policy that acts only on a region close to the goal, determined by the euclidean distance.
\cite{wang2020learning} builds upon Deep Deterministic Policy Gradients (DDPG) to incorporate existing base controllers into stages of exploration, value learning, and policy update, and present a straightforward way of synthesizing different base controllers to integrate their strengths.
\cite{nematollahi2021robot} proposes Soft-Actor Critic Gaussian Mixture Model (SAC-GMM), which is a hybrid approach that learns robot skills through a dynamical system modeled in state-space with GMMs and adapts the learned skills through interactions with the environment using RRL.
They present results in simulation for peg-insertion and power-lever-sliding skills, and real-world results for a door-opening task, using a camera image as a policy input.
\cite{spector2021insertionnet} propose the InsertionNet, which uses visual and wrench inputs to learn a residual policy in position and orientation.
Demonstrations are provided with a carefully designed procedure with backwards learning by first physically moving the robot to the final pose, e.g. the hole in the peg-in-hole task, and then generating collisions in order to collect data.
Data augmentation of images is used for robustness.
Having this dataset the insertion problem resorts to learning function parameters in a regression task.
To reach the insertion area they use a PD controller to follow a pre-computed computed trajectory.

The closest work to ours is the combination of Dynamic Movement Primitives (DMPs) with residual learning as done in~\cite{davchev2022residual}, where it was shown that RRL is better than learning the DMP parameters with RL.
Additionally, it is stated that learning orientation is important for reliable insertion tasks (several other works learn only the deviation in position).
Our work differs from~\cite{davchev2022residual} in that we make use of the variance in the demonstrations to check if the current position is inside the confidence interval and decide if another demonstration is needed, and when to adapt the nominal controller, while~\cite{davchev2022residual} specifies a time period after which the RL policy contributes to solving the task.
In~\cite{Cho2020Learning} insertion tasks have been learned with DMPs by tuning the parameters with episodic RL using Policy learning by Weighting Exploration with the Returns (PoWER)~\cite{Kober2011ML}, but the trajectories are demonstrated with kinesthetic teaching, which facilitates learning in joint space.
On the contrary, we work in cartesian space, which is inherently more difficult.

\section{Residual RL for Object-Centric ProMPs} \label{sec:approach}

In this section, we explain the different components of our proposed approach to combine ProMPs and RRL. An overview of the resulting method is shown in Fig.~\ref{fig:method}. We discuss the used underlying control structure in Section \ref{subsec:cart_control} and provide a short recap on ProMPs in Section \ref{subsec:promps} and on RL and SAC in Section \ref{subsec:rl}. Afterwards, we introduce our approach for adapting Cartesian ProMPs with Residual Robot Learning in Section \ref{subsec:rrl_promps} and explain the used action space and policy parametrization in Section \ref{subsec:actions_policy}.

\subsection{Cartesian Impedance Control}\label{subsec:cart_control}
When performing contact-rich tasks, such as an insertion, cartesian impedance control~\cite{Ott2003CartesianImpedance} is an appropriate choice, because it allows to specify the desired compliant behavior of the robot in presence of external forces, e.g. collisions.
This aspect is particularly relevant not only to prevent damaging the robot but also in human-robot collaboration.
Given a desired set-point (with zero velocity) of the end-effector pose $\eepose_{\text{des}} \in \R^7$, with position $\position_{\text{des}} \in \R^3$ and orientation represented as quaternion $\orientation_{\text{des}} \in \R^4$, the torques applied at the robot joints are computed as
\begin{align*}
  \torquevec_{\text{robot}} & = J(\jointpositionvec)^{\transpose} \wrench + C(\jointpositionvec, \jointvelocityvec)\jointvelocityvec + g(\jointpositionvec) \\
  \wrench & = \begin{bmatrix}
                \mat{K}_p (\position_{\text{des}} - \position) \\
                \mat{K}_q (\orientation_{\text{des}} \ominus \orientation)
              \end{bmatrix} + \mat{D}_d J(\jointpositionvec) \jointvelocityvec,
\end{align*}
where $\jointvelocityvec, \jointpositionvec \in \R^n$ are the current robot joint positions and velocities, $J(\jointpositionvec) \in \R^{6 \times n}$ is the Jacobian matrix, $C(\jointpositionvec, \jointvelocityvec) \in \R^{n \times n}$ and $g(\jointpositionvec) \in R^{n \times n}$ the coriolis forces and gravity compensation terms,
$\wrench \in \R^6$ is the simulated external wrench,
$\mat{K}_p \in \R^{3 \times 3} $ and $\mat{K}_q \in \R^{3 \times 3}$ the position and orientation stiffness matrices, $\mat{D}_d$ the damping matrix and $\ominus$ denotes a difference in quaternion space and translated to axis-angle.
Lower the gains $\mat{K}$ and $\mat{D}$ allows for safer robot interaction and exploration, but results in a larger tracking error.

\subsection{Probabilistic Movement Primitives}
\label{subsec:promps}
\glspl{mp} are a convenient way to represent time-based smooth robot and object movements~\cite{IjspeertDMP}.
In particular, probabilistic formulations allow to also capture variance in the demonstrations~\cite{calinon2007,Huang2019,paraschos2013promp}.
Here, we use ProMPs which are able to construct distributions that are conditioned on arbitrary time-steps and points inside a confidence interval of the demonstrations while relying on a small amount of training data, when compared to other state-based representations~\cite{2020iflowurain}.

Formally, a ProMP is a compact representation of a trajectory, where a point $\vec{y}_z \in \R^d$ in the trajectory is assumed to be a linear combination of $N$ basis functions $\vec{y}_z = \BasisFunctionMatrix_z^\transpose \prompweights$, with $\BasisFunctionMatrix \in \R^N$ a basis function matrix, $\prompweights \in \R^{N \times d}$ the learnable weights and $z(t) \in [0, 1]$ a phase-variable.
A distribution $p(\prompweights)$ over the weights is learned from multiple demonstrations. 
Assuming the weights are Gaussian distributed $p(\prompweights) = \mathcal{N}\left(\prompweights; \meanprompweights, \covprompweights \right)$, the mean and covariance matrix are obtained via maximum likelihood estimation.
For more details on the exact training procedure, we refer the reader to~\cite{paraschos2013promp,GomNeuSchPet20}.
Let $\overline{\vec{y}}_z$ be a point to reach at step $z(t)$ with covariance $\overline{\mat{\Sigma}}_z$. 
The conditional distribution over weights is computed with Bayes' rule for Gaussian distributions as $p(\prompweights | \overline{\vec{y}}_z, \overline{\mat{\Sigma}}_y) \propto p(\overline{\vec{y}}_z| \prompweights, \overline{\mat{\Sigma}}_y ) p(\prompweights) = \mathcal{N}\left( \prompweights; \meanprompweightsconditioned, \covprompweightsconditioned \right) $, and the resulting trajectory distribution $p(\overline{\trajectoryvec}) = \mathcal{N}\left( \overline{\trajectoryvec}; \BasisFunctionMatrix^\transpose \meanprompweightsconditioned, \BasisFunctionMatrix^\transpose \covprompweightsconditioned \BasisFunctionMatrix \right)$, with
\begin{align*}
    \meanprompweightsconditioned & = \meanprompweights^* + \mat{K} \left( \overline{\vec{y}}_z - \BasisFunctionVector_z^\transpose \meanprompweights^* \right), \quad \covprompweightsconditioned = \covprompweights^* - \mat{K} \BasisFunctionVector_z^\transpose \covprompweights^* \\
    \mat{K} & = \covprompweights^* \BasisFunctionVector_z \left( \overline{\mat{\Sigma}}_y + \BasisFunctionVector_z^\transpose \mat{\covprompweights^*} \BasisFunctionVector_z \right)^{-1} , \quad \BasisFunctionMatrix_z  = \mat{I}_d \otimes \BasisFunctionVector_z.
\end{align*}

\subsection{Reinforcement Learning and Soft-Actor Critic}\label{subsec:rl}

Let a \gls{mdp} be defined as a tuple $\mathcal{M} = (\statespace, \actionspace, \mathcal{R}, \mathcal{P}, \gamma, \mu_0)$, where $\statespace$ is a continuous state space $\state \in \statespace$, $\actionspace$ is a continuous action space $\action \in \actionspace$, $\mathcal{P}: \statespace \times \actionspace \times \statespace \to \R$ is a transition probability function, with $\mathcal{P}(\nextstate | \state, \action)$ the density of landing in state $\nextstate$ when taking action $\action$ in state $\state$, $\mathcal{R}: \statespace \times \actionspace \to \R$ is a reward function, $\gamma \in [0, 1) $ is a discount factor, and $\mu_0:\statespace \to \R$ the initial state distribution.
A policy $\policy(\action|\state)$ is a (stochastic) mapping from states to actions.
The state-action value function -- $Q$-function -- is the discounted sum of rewards collected from a given state-action pair following the policy $\policy$, ${Q^{\policy}(\state, \action) = \E{\policy,\Ptrans }{\sum_{t=0}^\infty \gamma^t r(\state_t, \action_t) | \state_0 = \state, \action_0 = \action}}$.
In general, the goal of a \gls{rl} agent is to maximize the expected sum of discounted rewards
${J(\policy) = \E{\tau \sim \mu_0, \policy, \mathcal{P}}{\sum_{t=0}^{\infty} \gamma^t r_t}}$.
In high-dimensional and continuous action spaces, typically a policy with parameters $\policyparams$ is updated iteratively with a gradient ascent step on $J$, using a variation of the policy gradient theorem~\cite{sutton1999PG}.
The \gls{sac} algorithm is a sample-efficient method to compute an off-policy gradient estimate of $\gradpolicyparams J(\policyparametrized)$~\cite{Degris2012OffPAC,Haarnoja2018SAC}, with several improvements over previous approaches, namely the entropy regularization and the squashed Gaussian policy. 
The entropy term encourages exploration by preventing the policy from becoming too deterministic during learning.
The surrogate objective optimized by \gls{sac} is
\begin{align*}
    J(\policyparams) = \E{\state \sim d^{\beta}, \action \sim \policytheta(\cdot | \state)}{Q^{\policy}_{\phivec}(\state, \action) -\alpha \log \policytheta (\action | \state)  },
\end{align*}
where $d^{\beta}$ is an off-policy state distribution, the $Q$-function is a neural network parameterized by $\phivec$ and $\alpha$ weighs the entropy regularization term.
The (unbiased) policy gradient of $J$ is computed by sampling from a replay buffer containing off-policy samples and using the reparametrization trick~\cite{kingma2014autoencoding} to differentiate the expectation over actions.

\subsection{Adapting Cartesian ProMPs with Residual Robot Learning}\label{subsec:rrl_promps}

\begin{figure*}[t]
  \includegraphics[width=\textwidth]{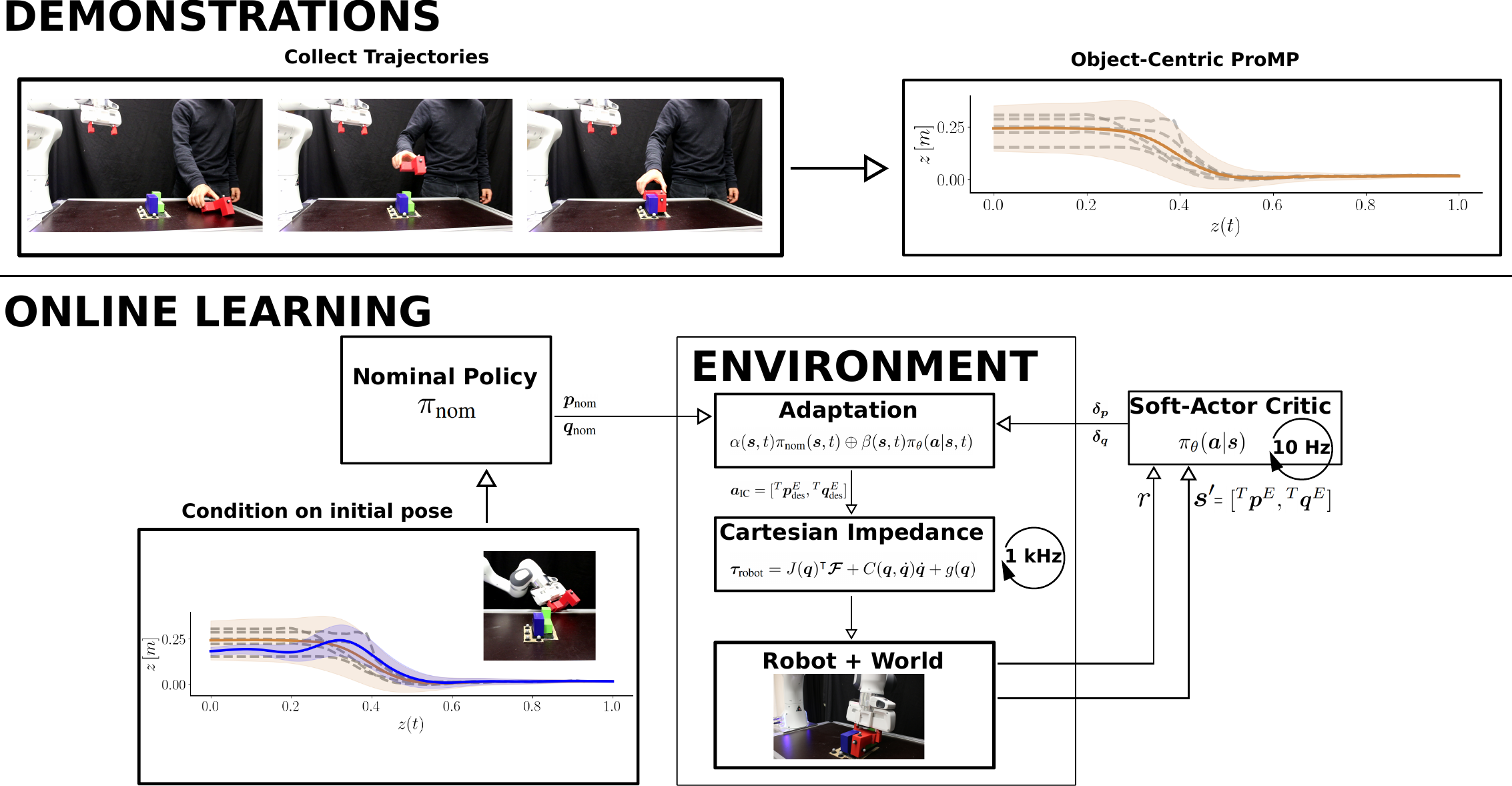}
  \caption{Our approach uses ProMPs to learn an object-centric trajectory generator from few demonstrations.
  At a new position within the confidence interval of the demonstrated trajectory distribution, the ProMP is conditioned and the mean trajectory is used to guide the robot to the goal area. 
  A residual policy learns deltas in position and orientation with SAC and an adaptation policy combines the action from $\policynominal$ with the ones from $\policytheta$.
  A low level cartesian controller implements the interface to the robot torques.
  }
  \label{fig:method}
\end{figure*}

Residual learning is commonly formulated as a combination of policies, which can be both time and state dependent as
\begin{align*}
    \policy(\action | \state, t) & = \Psi\left(\policynominal, \policytheta, \state, \action, t \right) \\
    & = \alpha(\state, t)  \policynominal(\state, t) \oplus \beta(\state, t) \policytheta(\action | \state, t),
\end{align*}
where $\policynominal$ is a nominal (or model-based) policy, $\policytheta$ is a learnable policy, and $\alpha$ and $\beta$ we call adaptation parameters. 
The operation $\oplus$ is dependent on the action space.
If $\action$ represents a translation, then it can be a sum, but if it is an orientation it can be a quaternion multiplication.
In~\cite{silver2018residual}, the authors assume $\policy(\state)=\policynominal(\state)+\policytheta(\state)$, and hence $\gradpolicyparams \policy(\state) = \gradpolicyparams \policytheta(\state)$, meaning one can use the policy gradient to optimize $\policyparams$ without knowing $\policynominal$.
However, this is equivalent to write the transition function of the residual MDP with the policy transformation $\mathcal{T}^{\policynominal}(\state, \action, \nextstate) = \mathcal{T}(\state, \Psi\left(\policynominal, \policytheta, \state, \action, t \right), \nextstate)$.
Because the transformation is now part of the environment, the agent is unaware of it.
Note that in the original formulation of~\cite{silver2018residual} the policy is only state-dependent, and not time-dependent. 
We augment this definition by using a time dependency, to include policies that result from time-dependent movement primitives such as ProMPs. This does not break the MDP assumption, since $t$ can be seen as part of the state (in an episodic task).

In our approach, we learn ProMPs from external observation of demonstrated object trajectories in cartesian space. For experiments in this paper, we used a motion capturing system and markers on the objects to obtain the demonstrated trajectories. An example of such a demonstration can be seen in Fig. \ref{fig:method} on the top left.
We assume that for an insertion task we collect a set of $N$ trajectories $\{\trajectoryvec_i \}_{i=1,\ldots,N}$ of the pose of an \textit{interest} object \textit{I} (red object in Fig.~\ref{fig:real_robot_ubongo}) in the reference frame of a \textit{target} object \textit{T} (blue object), where $\trajectoryvec_i = \left[\interestintargetpose_0, \ldots, \interestintargetpose_{H_i-1} \right]$, with $H_i$ the $i$-th trajectory length, and $\interestintargetpose = (\position, \orientation)$, with $\position \in \R^3$ is the position and $\orientation \in \R^4$ the orientation (as quaternion). 
Representing the trajectory in the target frame allows to get the intention of the demonstration, i.e. classify if it is an insertion task, and additionally, if the target or interest objects move to a different pose, their relation is maintained.
For a compact representation of the trajectory, we encode the position with an object-centric ProMP in cartesian space.
The orientation representation is the average over trajectories.
Learning ProMPs for orientation spaces in an ongoing topic of research~\cite{2021RozoOrientationpromp}, and we leave this for future work.

A key advantage of movement primitives is their ability to generalize from demonstrated trajectories to new situations. In particular, here ProMPs can be used to compute nominal trajectories for varying start positions of the object. In particular, they can also distinguish between starting points covered by the provided demonstrations and starting points outside this region. We focus here on generalization to starting points within the demonstrated distribution over trajectories only. However, as a future direction it would be also possible to include active requests for additional demonstrations or multi-modal ProMPs using incremental Gaussian Mixture Models~\cite{2019koertLearningintention}.

For our approach we compute the nominal trajectory by conditioning the ProMP on the initial time-step $z=0$ and initial pose $\overline{\vec{y}}_0 \equiv \interestintargetpose$ with a small covariance $\overline{\mat{\Sigma}}_0$ and compute the resulting mean trajectory $\overline{\trajectoryvec} = \BasisFunctionMatrix^\transpose \meanprompweightsconditioned$. 
Afterwards, the desired trajectory $\overline{\trajectoryvec}$ is translated to the end-effector in the target frame with ${\eeintargetpose = \interestintargetpose \eeininterestpose}$, where a transformation from the end-effector to the interest object is given via a grasping pose.

To follow $\overline{\trajectoryvec}$ we could compute the inverse kinematics and track it with an inverse dynamics controller in joint space.
However, this strategy would require a large gain to ensure a low tracking error, necessary for an insertion task with limited tolerance.
These large gains could damage the robot and the environment in case of interaction.
For this reason, a low gain controller in cartesian space is better suited for this task.
However, due to the low gains, velocity constraints, among others, the controller will not perfectly follow the desired trajectory and thus cannot complete the task, as depicted in the experiment of Fig.~\ref{fig:snapshot_nominal}.

An important decision is where/when to activate the residual part of the policy.
In the original formulation of residual RL $\alpha=\beta=1$, which for longer trajectories can lead to exploring in free-space regions far away from the insertion goal.
To prevent this unnecessary exploration, \cite{lee2020guided} and \cite{shi2021combining} set $\alpha(\state, t) = 0,\, \beta(\state, t) = 1$ if $\state \in \mathcal{S}_u$, a region in the vicinity of the goal, and $\alpha=1,\, \beta=0$ otherwise.
\cite{lee2020guided} defines $\mathcal{S}_u$ based on an uncertainty quantification in pose estimation, and \cite{shi2021combining} defines $\mathcal{S}_u$ as the distance to the goal, both being hyperparameters.
\cite{kulkarni2021learning} uses a time-based weighting as $\alpha(t)=1 - \beta(t)$, with $\beta(t) = \max(0, (T-t)/T)$, with $T$ the settling time of the nominal controller, which in practice means that the learned policy only acts if the nominal controller fails, and the residual part of the policy acts alone in the environment.
In~\cite{davchev2022residual} $\alpha=\beta=1$ after executing the nominal controller for a certain amount of time (in their task $3.9$ sec), and otherwise $(\alpha, \beta)=(1, 0)$.

In our approach, we make use of the covariance over originally demonstrated trajectories computed with $\BasisFunctionMatrix^\transpose \covprompweights$.
For our particular insertion task, we motivate this with the intuition that exploration is more beneficial closer to the insertion location, that have lower entropy (less variance), as can be seen from the last time steps of the learned ProMPs in Fig.~\ref{fig:promps}.
We propose a \textit{variance-based adaptation} scheme with $\alpha=1$ and
\begin{align}
    \beta(\state, t) = \begin{cases}
    1 & \text{if } \exists i: \min_i{\sigma}_{i}(t) \leq \varepsilon \\
    0 & \text{otherwise}
    \end{cases},
    \label{eq:variance-based-adaptation}
\end{align}
where $\sigma_{i}(t)$ is the standard deviation of the $i$-th dimension of the original ProMP at time step $t$.

\subsection{Action Space and Policy Parametrization}\label{subsec:actions_policy}

For the RL agent, the cartesian impedance controller is part of the environment and takes as input a desired pose ${\action_{\text{IC}} =  [\eeintargetpos_{\text{des}}, \eeintargetori_{\text{des}} ]}$, computed with the adaptation scheme.
The policy $\policytheta$ is learned with SAC, and encodes the mean and variance of a Gaussian distribution over positions and orientations ${\policytheta(\action | \state) = \mathcal{N}\left(\deltaposition; \mu_{\thetavec}^{\position}(\state), \Sigma_{\thetavec}^{\position}(\state) \right) \mathcal{N}\left(\deltaaxisangle; \mu_{\thetavec}^{\axisangle}(\state), \Sigma_{\thetavec}^{\axisangle}(\state) \right) }$.
The functions encoding the mean and covariance are neural networks that share the same features up to the last linear layer.
The delta in orientations $\deltaaxisangle \in \R^{3}$ parametrizes the coordinates of an axis-angle representation.
Both positions and orientations are squashed to the cartesian controller limits with a hyperbolic tangent operator.
The nominal and learned policies are combined as follows.
For the position it is a simple addition $ \eeintargetpos_{\text{des}} = \position_{\text{nom}} + \deltaposition$.
For orientations, we first compute the quaternion representation of $\deltaaxisangle$ as ${\deltaquaternion = [\cos(\|\deltaaxisangle\|/2), \deltaaxisangle/\|\deltaaxisangle\| \sin(\|\deltaaxisangle\| / 2)]}$, and afterwards apply a quaternion multiplication to obtain the desired rotation $ \eeintargetori_{\text{des}} = \deltaquaternion \circ \orientation_{\text{nom}}$.

\section{Experimental Results}
\label{sec:experiments}

\begin{figure}[t]
  \centering
  \includegraphics[width=\columnwidth]{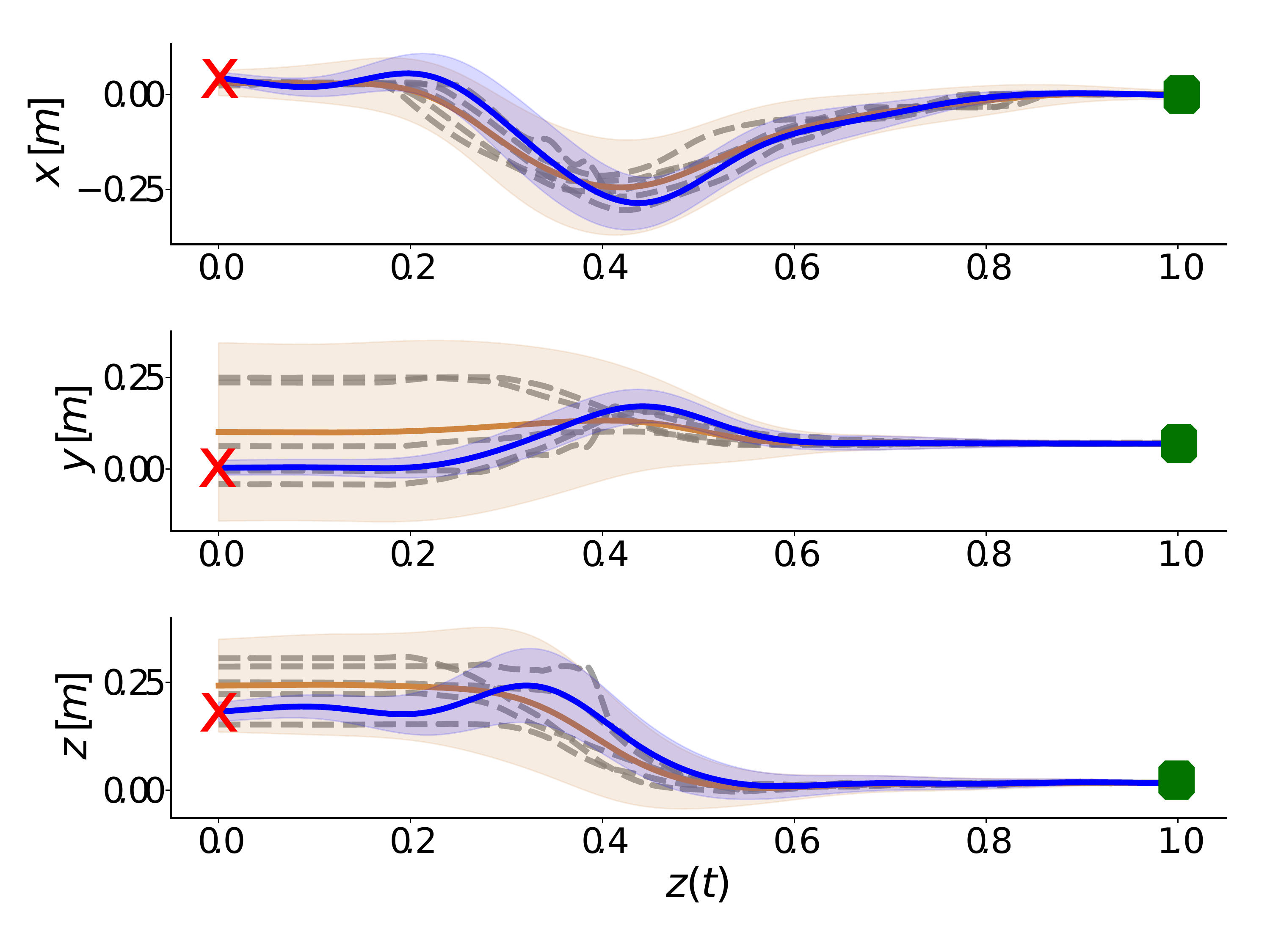}
  \\
  \includegraphics[width=0.75\columnwidth]{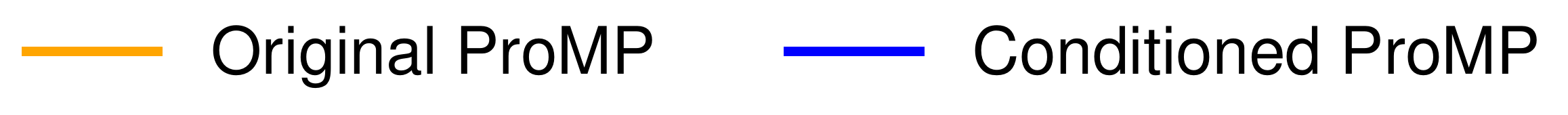}
  \caption{
  Human demonstrated trajectories (gray lines) and ProMPs of interest object position in the target object frame. The overall ProMP learned from demonstrations is shown in orange and the result after conditioning is shown in blue.
  Solid lines represent the mean and the shaded area two times standard deviations of the ProMPs.
  The red cross indicates the conditioned observation at time $z=0$, and the green square the trajectory goal at $z=1$.
  }\
  \label{fig:promps}
\end{figure}

\begin{figure*}[t]
  \begin{minipage}[c]{0.19\linewidth}
    \includegraphics[width=\textwidth]{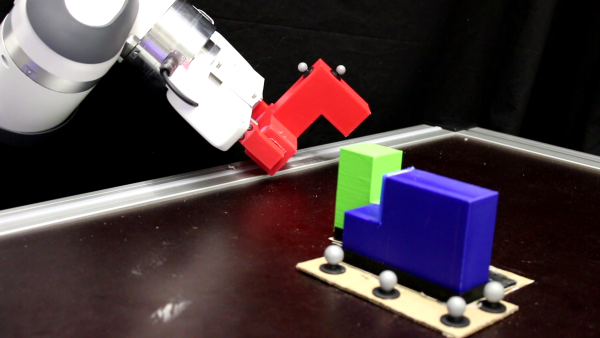}
  \end{minipage}
  \hfill
  \begin{minipage}[c]{0.19\linewidth}
    \includegraphics[width=\textwidth]{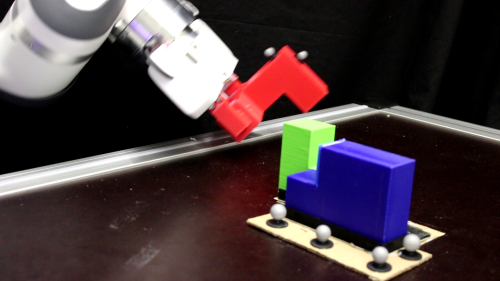}
  \end{minipage}  
  \hfill
  \begin{minipage}[c]{0.19\linewidth}
    \includegraphics[width=\textwidth]{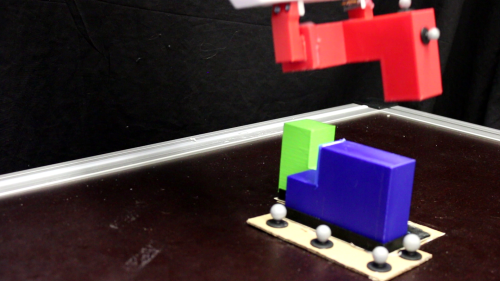}
  \end{minipage}  
  \hfill
  \begin{minipage}[c]{0.19\linewidth}
    \includegraphics[width=\textwidth]{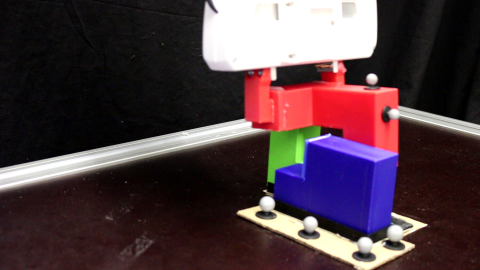}
  \end{minipage}  
  \hfill
  \begin{minipage}[c]{0.19\linewidth}
    \includegraphics[width=\textwidth]{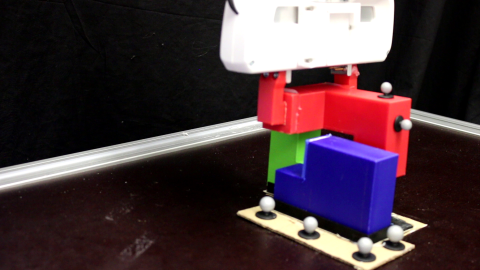}
  \end{minipage}  
  
  \caption{Snapshot an episode using the nominal controller only.
  This policy fails to precisely track the trajectory, due to the low gains used to ensure safe interaction.
  }
  \label{fig:snapshot_nominal}
\end{figure*}

\begin{figure*}[t]
  \begin{minipage}[c]{0.19\linewidth}
    \includegraphics[width=\textwidth]{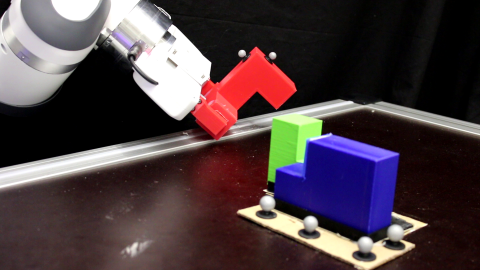}
  \end{minipage}
  \hfill
  \begin{minipage}[c]{0.19\linewidth}
    \includegraphics[width=\textwidth]{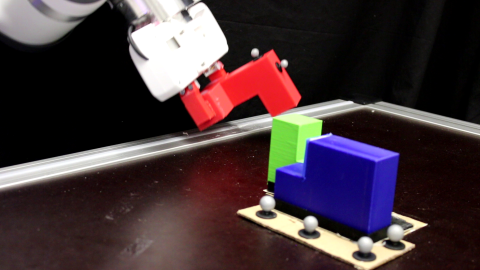}
  \end{minipage}  
  \hfill
  \begin{minipage}[c]{0.19\linewidth}
    \includegraphics[width=\textwidth]{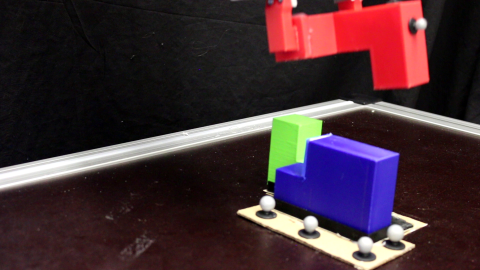}
  \end{minipage}  
  \hfill
  \begin{minipage}[c]{0.19\linewidth}
    \includegraphics[width=\textwidth]{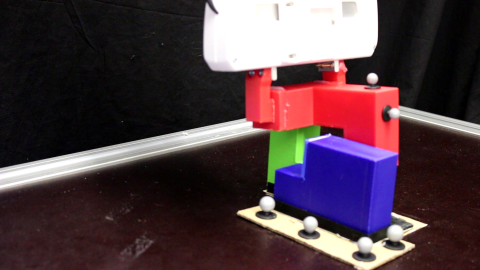}
  \end{minipage}  
  \hfill
  \begin{minipage}[c]{0.19\linewidth}
    \includegraphics[width=\textwidth]{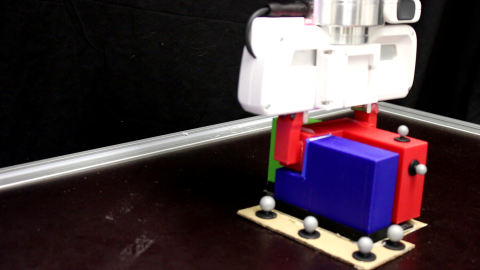}
  \end{minipage}  
  
  \caption{Snapshot of an episode using the learned policy of Residual RL with variance-based adaptation.
  By using the residual in low entropy regions of the state-space, such as ones close to the insertion point, the RL agent only needs to learn to adapt closer to the goal.
  Notice how the $4$th image closely matches the one from Figure~\ref{fig:snapshot_nominal}, showing that the agent follows the nominal policy and only adapts near the goal.
  }
  \label{fig:snapshot_variance}
\end{figure*}

\begin{figure*}[t]
  \includegraphics[width=0.32\textwidth]{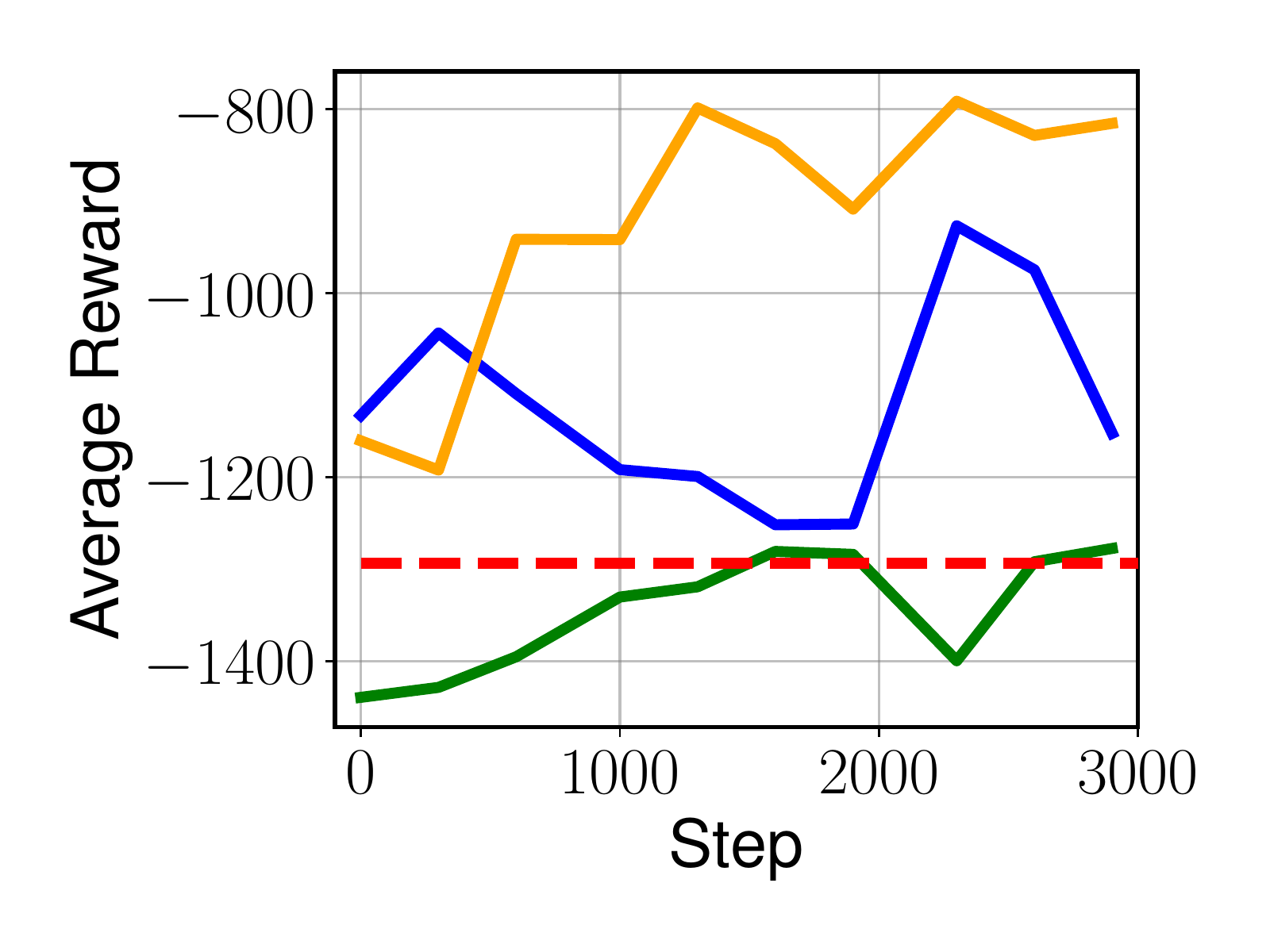}
  \hfill
  \includegraphics[width=0.32\textwidth]{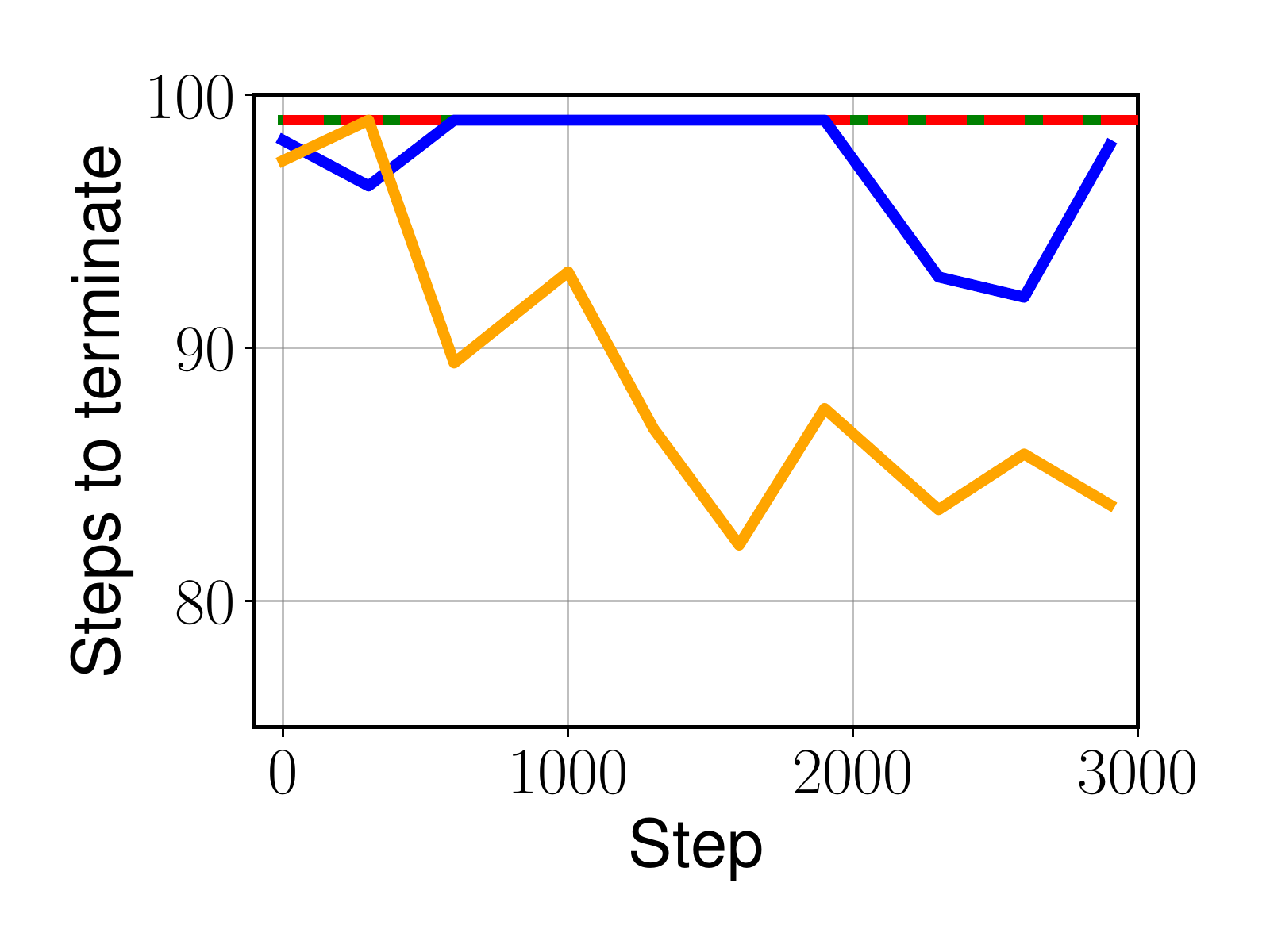}
  \hfill
  \includegraphics[width=0.32\textwidth]{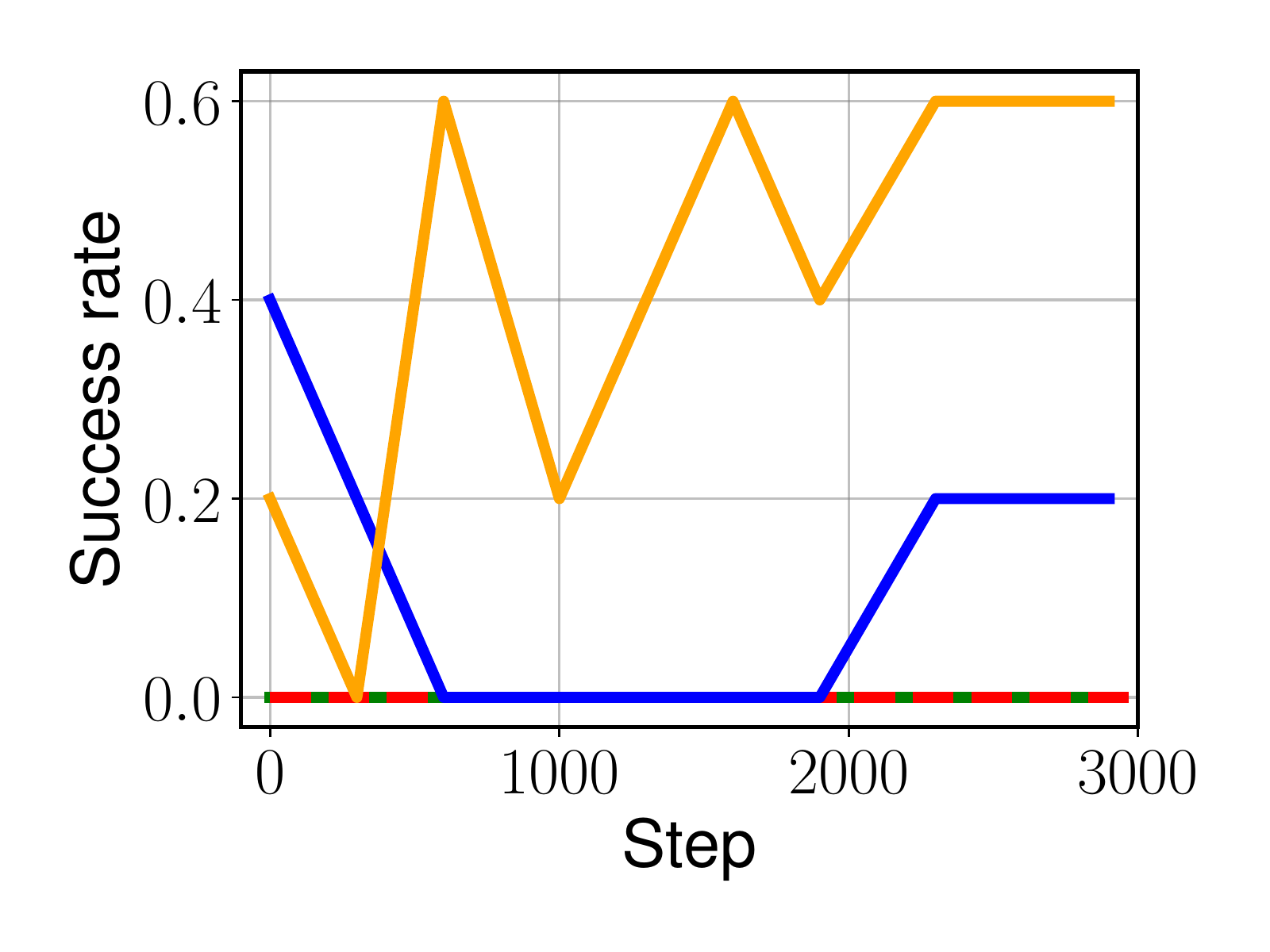}
  \\
  \includegraphics[width=\textwidth]{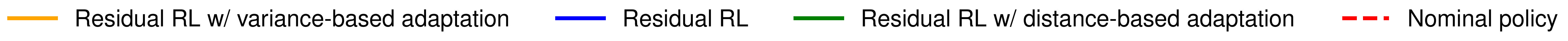}
  \caption{Real robot results. 
  The plots show a qualitative trend that residual model-free RL is able to adapt ProMPs for the Ubongo$3$D task.
  The solid lines depict the mean over 5 trials with different random seeds in the real system.
  }
  \label{fig:experiment_results}
\end{figure*}

As a proof of concept, we evaluate our proposed method in a $3D$ block insertion task with a 7-DoF Franka Panda Robot. In these experiments, we investigate if the task, which we could not solve with basic ProMPs before, benefits from the combination of ProMPs and RRL and compare different ways of combining nominal and residual policy to assess which one works best in the real system.

\subsection{Experiment Setup}
We evaluate the proposed method on an approximation of the Ubongo3D game~\cite{ubongo3d}, which consists of different shapes that have to be assembled together in a limited space. For the experiments in this paper, we use 3 different shapes - the red, green and blue elements in Fig.~\ref{fig:real_robot_ubongo} - and a base plate (in black).
The tolerance for insertion is approximately $3$mm.
Each shape is a custom $3$D printed structure, consisting of cubes with $3.5$cm sides.
The goal is to build a \textit{a priori} unknown structure with a height of two cubes and such that the base plate is covered.
The game has a planning and a manipulation part.
Solving the planning problem involves deciding the pose of each shape and could e.g. be done using Mixed-Integer Programming~\cite{conforti2014IntegerProgramming}.
On the other hand, stacking/inserting the shapes together can be seen as a fine manipulation task, which is knowingly difficult for robots \cite{kroemer2021review}.

We see the Ubongo $3$D task as a proxy for more complex assembly scenarios with small tolerances. Due to the involvement of multiple points of contact for the insertion, we also consider it a particularly suitable task for model-free RL approaches, since they do not rely on accurate models which would be hard to obtain in practice. Additionally, the insertion here is non-invariant to changes in orientation and therefore requires learning orientations as part of the residual policy.

For the experiments in this paper, we assume the blue and green shapes are already placed and fixed and learn how to insert the red shape from human demonstrated trajectories.
Therefore, demonstrations are recorded starting from different initial configurations, using a motion capturing system and markers attached to the objects. We record the positions and orientations of the interest object pose (the red shape) in the reference frame of a target object (the blue shape). It is notable here that these demonstrations are recorded from external observations in Cartesian space, i.e. a human demonstrator moving the objects and not with kinesthetic teaching on the robot.
Fig.~\ref{fig:promps} shows the $5$ recorded trajectories, the resulting learned ProMP (orange), and an example for conditioning on a new (not initially demonstrated) initial position (blue). For learning the ProMP we decided to use $10$ basis function after comparing the data log-likelihood in a grid search.

The cartesian impedance controller runs at $1000$Hz and learnable policy at $10$Hz.
The task is executed episodically with $100$ steps, amounting to approximately $15$ seconds per episode, and a learning trial took approximately $15$ minutes.
The episode terminates if the interest object (red shape) is at a distance of the goal position and orientation less than
$5$mm and  $5^{\circ}$, respectively.
The state is the position and orientation (as quaternion) of the end-effector in the target object $\state = [\eeintargetpos, \eeintargetori] \in \R^7$.
While recent works~\cite{kulkarni2021learning} also use the external wrench expressed in the target frame, we found in our experiments that the measurement provided by the Panda robot was too unreliable and not useful for learning.

We experimented using a sparse reward, but found that the task was too difficult to learn without a reward signal. 
We hypothesize that this was due to the termination condition being too strict.
Quite often the red shape was already fairly in place, but not enough to get to terminate and obtain a reward.
For a peg-in-hole task, many works use a sparse reward, but once the peg is inside the hole the region of exploration is lower, especially with a fixed orientation and the agent simply has to push down the peg.
In our case, we have a combination of an insertion and precise placing task, where after starting the insertion, the red shape can still rotate around one axis.
Since an approximation of the final pose is known, we decided to use instead a per-step dense reward function that weighs the absolute distances of the error in position and orientation $r(\state, \action) = -\sum_{i=1}^3|\eeintargetpos_{\text{des}} - \eeintargetpos_i| - 50 \|(\orientation_{\text{des}} \ominus \orientation)\|$.
Using dense rewards is also common in other works that learn in the real system~\cite{johannink2019residual}.
The discount factor was $\gamma=0.99$.

The policy and $Q$-functions are two-layer neural networks with $128$ hidden neurons and ReLU activations,
and the entropy regularization weight in SAC ($\alpha$) is also learned.
All parameters are optimized with ADAM~\cite{kingma2014method} and a learning rate $3 \cdot 10^{-4}$. 
The initial replay size and the number of samples before policy updates is $100$, as well as the batch size for actor and critic learning.
In practice, we did not find the need to use a recurrent policy as in~\cite{kulkarni2021learning,davchev2022residual}.

\subsection{Results}

We report the results of executing the nominal controller and three adaptation strategies: 
\begin{itemize}
    \item \textit{Residual RL} - the residual policy is always active
    \item \textit{Residual RL with variance-based adaptation} (ours) - detailed in Eq.~\ref{eq:variance-based-adaptation}
    \item \textit{Residual RL with distance-based adaptation} - if the distance to the goal is less than a threshold, use only the residual policy as in~\cite{shi2021combining}. The threshold was chosen empirically such that it occurs almost before insertion as in the $4$-th picture in Fig.~\ref{fig:snapshot_nominal}.
\end{itemize}

Figures~\ref{fig:snapshot_nominal} and~\ref{fig:snapshot_variance} show snapshots of the execution of the nominal controller and the learned residual policy with the variance-based adaptation strategy, respectively.
Notice that the initial position starts far away from the goal, which makes it more difficult to precisely track the object trajectory until the insertion point.

The average reward, number of time steps to succeed, and the success rate during training of the three adaptation strategies are depicted in Fig.~\ref{fig:experiment_results}.
From the figures, we can observe that all methods show an improvement trend on the average reward curve (most left plot).
The results show that while the nominal controller (red dashed line) cannot solve the task due to the low-gain controller, our method (yellow line), and residual RL (blue line) can improve it, as seen in the average reward plot. 
Using the residual RL policy for the whole trajectory can lead to explore far from the insertion point and small deviations can accumulate over time, leading to a point where the red shape gets stuck and cannot finish the insertion.
In the rightmost plot, the success rate at step $0$ is different from $0$, because some trials could already perform an insertion just by adding small Gaussian noise around the nominal trajectory.
Even though we initialize the policy neural network to output means close to $0$, as is common in RRL, by construction the network computing the variance outputs a value different from $0$, making it sufficient to move the shape slightly around the nominal trajectory.
For the distance-based adaptation (green lines) we made sure to only switch fully to the learnable controller when the red shape is already in contact with the green shape ($4$th frame in Fig.~\ref{fig:snapshot_nominal}).
While this strategy is slowly learning - note the increase in reward the leftmost plot of Fig.~\ref{fig:experiment_results} - meaning the policy is successfully bringing the red shape to the goal location, it could not complete the task in any of the trials, as was expected due to the random exploration.
Lastly, the steps it takes for the episode to terminate are correlated with the success rate.

\section{Conclusion and Future Work} \label{sec:conclusion}
In this paper, we studied how to use human demonstrations of object-centric trajectories in combination with ProMPs and residual learning.
Because the cartesian impedance controller has a low gain to guarantee safe interaction, simply following the trajectory from a conditioned ProMP resulted in task failure.
We overcome this problem by proposing to learn a residual policy in position and orientation with model-free reinforcement learning.
Making use of the variability in the demonstrations, we present an adaptation strategy based on the variance of the ProMP as an indication of a region where an insertion task can take place, and thus where the policy needs to be learned, thereby increasing the sample efficiency.
The experimental evaluations in the real robot showed that our method is able to learn a policy that corrects the ProMP trajectory to perform a block insertion task in the Ubongo$3$D game.

In future work we plan to evaluate our approach in other tasks with different objects, include a formulation of orientation ProMPs in Riemannian Manifolds~\cite{Rozo&Dave*2021}, learn the adaptation strategy parameters as part of the policy learning, and study other trajectory representation methods using state-space information~\cite{Khansari-Zadeh:174759,2020iflowurain}.

\bibliographystyle{IEEEtran}
\bibliography{references}

\end{document}